\documentclass[letterpaper, 10 pt, conference]{ieeeconf}  

\IEEEoverridecommandlockouts                              

\usepackage{cite}
\usepackage{amsmath,amssymb,amsfonts}
\usepackage{multirow}
\usepackage{algorithmic}
\usepackage{graphicx}
\usepackage{textcomp}
\usepackage{xcolor}
\usepackage{booktabs}
\usepackage{makecell}
\def\BibTeX{{\rm B\kern-.05em{\sc i\kern-.025em b}\kern-.08em
    T\kern-.1667em\lower.7ex\hbox{E}\kern-.125emX}}
\begin{document}

\title{FastATDC: Fast Anomalous Trajectory Detection and Classification*\\
\thanks{*This work was supported in part by the National Natural Science Foundation of China under grants 61873191 and 72074170, and in part by the Science and Technology Commission of Shanghai Municipality under grant 20JG0500200. (\textit{Corresponding author: Jingwei Wang.})     \\
\indent Tianle Ni is with the College of Electrical Engineering and Information Technology, Technical University of Munich, Munich 80333, Germany (e-mail: tianle.ni@tum.de).  \\
\indent Jingwei Wang, Yunlong Ma, Min Liu are with the College of Electronic and Information Engineering, Tongji University, Shanghai 201804, China (e-mails: jwwang@tongji.edu.cn; evanma@tongji.edu.cn; lmin@tongji.edu.cn). \\
\indent Shuang Wang is with Shanghai Police College, Shanghai 200137, China (e-mail: shpdxqcjl@163.com). \\
\indent Weiming Shen is with the State Key Laboratory of Digital Manufacturing Equipment and Technology, Huazhong University of Science and Technology, Wuhan 430074, China (e-mail: wshen@ieee.org).}
}

\author{Tianle Ni, Jingwei Wang, Yunlong Ma, Shuang Wang, Min Liu, and Weiming Shen, \textit{Fellow, IEEE}}

\maketitle

\begin{abstract}
Automated detection of anomalous trajectories is an important problem with considerable applications in intelligent transportation systems. Many existing studies have focused on distinguishing anomalous trajectories from normal trajectories, ignoring the large differences between anomalous trajectories. A recent study has made great progress in identifying abnormal trajectory patterns and proposed a two-stage algorithm for anomalous trajectory detection and classification (ATDC). This algorithm has excellent performance but suffers from a few limitations, such as high time complexity and poor interpretation.
Here, we present a careful theoretical and empirical analysis of the ATDC algorithm, showing that the calculation of anomaly scores in both stages can be simplified, and that the second stage of the algorithm is much more important than the first stage. Hence, we develop a FastATDC algorithm that introduces a random sampling strategy in both stages.
Experimental results show that FastATDC is 10 to 20 times faster than ATDC on real datasets. Moreover, FastATDC outperforms the baseline algorithms and is comparable to the ATDC algorithm.
\end{abstract}


\section{Introduction}
Anomaly detection is a major research problem in unsupervised learning, with a wide range of important applications such as financial risk control \cite{du2021application,kamivsalic2021synergy}, industrial equipment maintenance \cite{huang2021digital}, traffic data mining \cite{ganapathy2021data,salazar2021traffic}. Anomalies usually exhibit specificity, randomness, and low-frequency characteristics, and the definition of anomalies varies for different tasks. These factors are what make anomaly detection so challenging.

In this paper, we focus on the automated detection of anomalous trajectories, which is an important component of intelligent transport systems (ITS)\cite{rudskoy2021digital}. An anomalous trajectory refers to a trajectory that differs locally or globally from most other normal ones when measured by some similarity metrics. Many ITS applications, such as fraud detection\cite{lakhan2022its}, emergency response \cite{audu2021application}, and urban network mapping \cite{autili2021cooperative}, have the urgent need for anomalous trajectory detection. Over the past decade, much research has been devoted to developing anomalous trajectory detection methods, such as density-based methods \cite{lee2008trajectory} and isolation-based methods \cite{Zhang2011iBAT, Chen2013iBOAT}. However, these studies only focus on distinguishing anomalous trajectories from normal ones, without considering that anomalous trajectories are also extremely different from each other. A recent study \cite{Wang2020ATDC} has noticed this and sorted out four different patterns of abnormal trajectories: global detour (GD), local detour (LD), global shortcut (GS), and local shortcut (LS). This study also proposed a new algorithm, called the anomalous trajectory detection and classification (ATDC) algorithm, which achieved excellent performance.

However, the ATDC algorithm suffers from some limitations. On the one hand, ATDC uses a two-stage strategy to calculate the anomaly score of trajectories, in which each stage traverses a large number of trajectories. As a result, the overall time complexity of the algorithm is $O(N^2)$, where $N$ is the number of trajectories. This makes the algorithm difficult to use on large-scale datasets or in online detection scenarios. On the other hand, despite the excellent performance of this two-stage algorithm, there is no in-depth analysis of why it works. For example, it is unclear exactly which of these stages plays the main role.

In this paper, we perform a careful analysis of the ATDC algorithm and find that each stage is equivalent to a $k$-nearest neighbor algorithm. The empirical results show that the second stage is significantly more important than the first one. Moreover, we theoretically prove that the calculation of the anomaly scores used in this algorithm can be simplified. Consequently, we develop a fast anomalous trajectory detection and classification (FastATDC) algorithm that reduces the time complexity to $O(kN)$ where $k\ll{N}$, by introducing a random sampling strategy. Experimental results show that FastATDC achieves comparable performance in less than a tenth of the computation time of ATDC. The main contributions are summarized as follows:
\begin{itemize}
\item We present a careful theoretical and empirical analysis of the ATDC algorithm to provide a better understanding of its mechanism and performance.
\item We propose the FastATDC algorithm that introduces a random sampling strategy in both stages of ATDC, allowing the time complexity to be reduced from $O(N^2)$ to $O(kN)$. 
\item Extensive experiments on real datasets show that the FastATDC algorithm excels the baselines and even outperforms ATDC in some cases. Moreover, FastATDC is robust to the sampling rates and is 10 to 20 times faster than ATDC.
\end{itemize}

\section{Related Work}

Anomalous trajectory detection is a challenging problem and a few classical algorithms have been proposed, such as TRAOD\cite{lee2008trajectory} and iBAT\cite{Zhang2011iBAT}. 
TRAOD is a two-phase method proposed by Lee et al., which first partitions a trajectory into many sub-trajectories and detects the anomalous ones from these sub-trajectories. Liu et al. further improved TRAOD and introduced a density-based trajectory outlier detection (DBTOD) algorithm to better distinguish anomalous sub-trajectories from normal ones\cite{6545447}. iBAT is a faster detection algorithm put forward by Zhang et al., which applied the isolation mechanism to detect taxi driving fraud. Based on iBAT, Chen et al. further designed the iBOAT algorithm, which is an online detection method that compares the unknown trajectories against normal ground truth ones via an adaptive working window \cite{Chen2013iBOAT}.
Furthermore, Wang et al. found that patterns of anomalous trajectories are not unique but multiple and uncertain, and applied an adaptive hierarchical clustering method \cite{wang2018detecting}. Zhu et al. utilized top-k most popular traces as a reference and used an Edit distance in both spatial and temporal domains to measure the difference between trajectories \cite{zhu2015time}. Inspired by these algorithms and findings, Wang et al. proposed a two-stage method called ATDC for anomalous trajectory detection and classification \cite{Wang2020ATDC}, which is the foundation of our work.


\section{Methodology}

\begin{figure}[t!]
\centering
\includegraphics[width=0.44\textwidth]{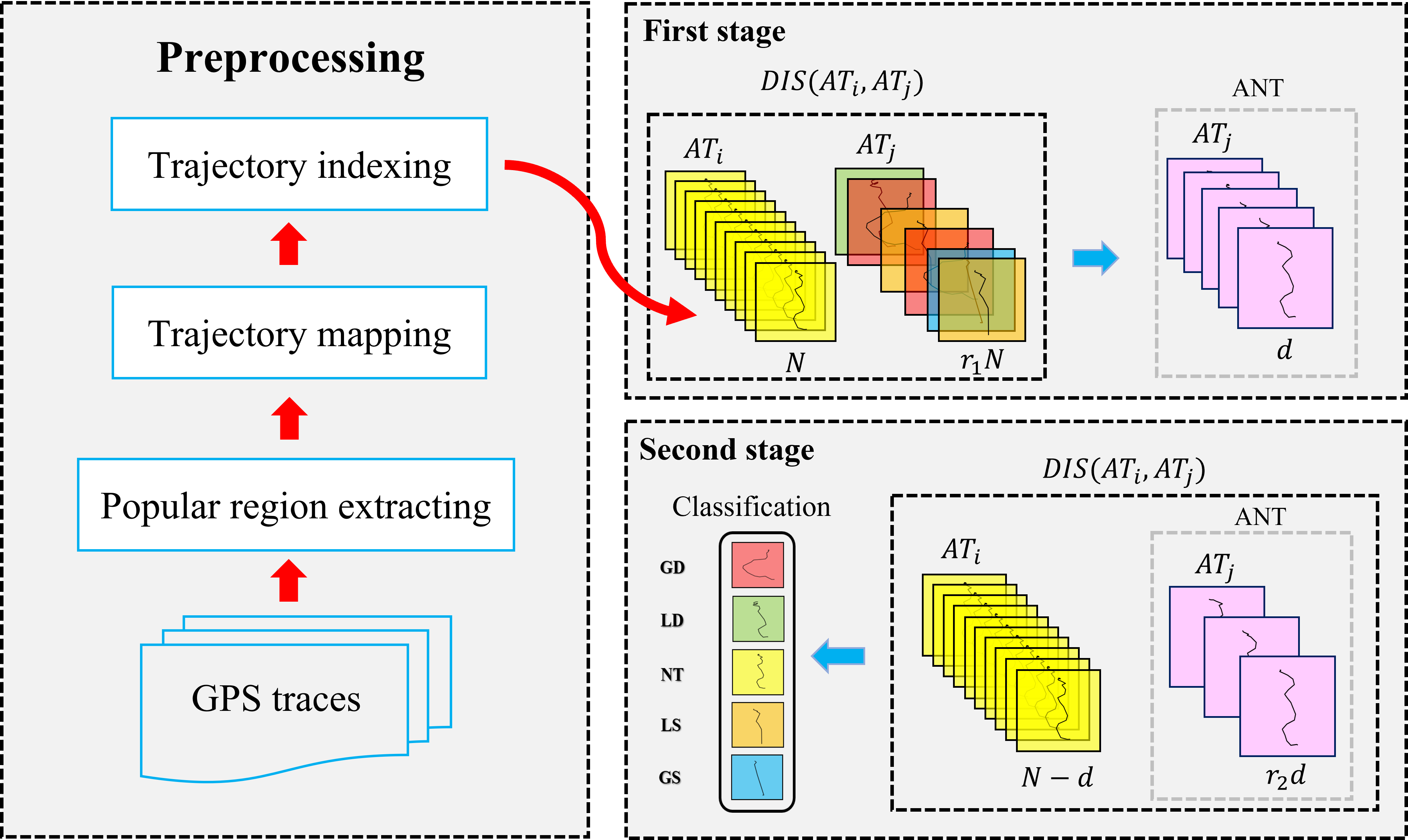}
\caption{\label{fig:ATDC}The framework of the FastATDC algorithm.}
\end{figure}

In this section, we first briefly introduce the ATDC algorithm and then provide a theoretical analysis of the trajectory anomaly score function in ATDC, followed by the proposed FastATDC algorithm, which is summarized in Fig. 1.

\subsection{ATDC}
Note that the preprocessing of trajectory data in our work is the same as in \cite{Wang2020ATDC}, and details (including concepts and notations) can be found in this paper and will not be repeated here due to the limited space.

Wang et al. proposed the difference and intersection distance (DIS) metric to quantify the difference between any two trajectories \cite{Wang2020ATDC}. Given a trajectory dataset with a set of trajectories $\tau = \{AT_1,AT_2,...,AT_N\}$, the DIS distance between two trajectories $AT_i$ and $AT_j$ is defined as
\begin{equation}
{DIS}(AT_i,AT_j) = \frac{|AT_i|-|AT_j|}{|AT_i \cap AT_j|}.
\end{equation}

ATDC consists of two stages, each of which calculates the anomaly score for each trajectory. At the first stage, the anomaly score of trajectory $AT_i$ is defined as
\begin{equation}
\label{S1}
{S_1}(AT_i) = \frac{\sum_{j,j\neq i}^{N-1} \Big(|AT_i|-|AT_j| \Big)}{\sum_{j,j\neq i}^{N-1} |AT_i \cap AT_j|}.
\end{equation}

After the first stage calculation, the trajectories with the most frequent patterns have the property that their anomaly scores ${S_1}$ are within a small interval $[-\phi,\phi]$. These trajectories are defined as absolute normal trajectories (ANT), denoted by $\tau_{0}$. Those trajectories that are not in the set $\tau_{0}$ will be sent to the next stage.
In the second stage, the anomaly score of trajectory $AT_i$ is calculated by 
\begin{equation}
\label{S2}
{S_2}(AT_i) = \frac{\sum_{j,j\in \tau_{ik}}^{k} \Big(|AT_i|-|AT_j|\Big)}{\sum_{j,j\in \tau_{ik}}^{k} |AT_i \cap AT_j|},
\end{equation}
where $\tau_{ik} \subset \tau_{0}$ is the set of $k$ closest trajectories of trajectory $AT_i$, defined as $k$ trajectories with the largest intersection with $AT_i$.

Taken together, the anomaly score of one trajectory $AT_i$ is defined as
\begin{equation}
{S}(AT_i) = 
\begin{cases}
{S_1}(AT_i), \quad &AT_i \in \tau_{0}    \\
{S_2}(AT_i), \quad &AT_i \notin \tau_{0}.
\end{cases}
\end{equation}

ATDC uses a segmentation function $L$ to distinguish between four types of abnormal trajectories (i.e., GD, LD, LS, GS) and normal trajectories (NT).
In particular, 
\begin{equation}
\label{L}
L(AT_i;\tau,\boldsymbol{\theta}) = 
\begin{cases}
AT_i \; is \; GD, & S(AT_i) \geq \theta_1  \\
AT_i \; is \; LD, & \theta_2 \leq S(AT_i) < \theta_1  \\
AT_i \; is \; NT, & \theta_3 < S(AT_i) < \theta_2     \\
AT_i \; is \; LS, & \theta_4 < S(AT_i) \leq \theta_3  \\
AT_i \; is \; GS, & S(AT_i) \leq \theta_4. \\
\end{cases}
\end{equation}
where $\boldsymbol{\theta}= [ \theta_1, \theta_2, \theta_3, \theta_4 ]$ is a pre-defined threshold vector.

\subsection{Analysis of ATDC}

Now we reconsider the anomaly score $S_1(AT_i)$ under the context of random sampling. Suppose a dataset contains five classes of trajectories $\{GD, LD, NT, LS, GS\}$, and $c$ represents one of the five classes. The probability of different trajectories being randomly drawn is $\mathbf{P}=(P_{GD},P_{LD},P_{NT},P_{LS},P_{GS})$, where $P_{NT}$ are much larger than others. Taking dataset T-1 as an example (see Table I), the probability vector is $\mathbf{P}=(1.5\%,2.2\%,86.9\%,9.1\%,0.3\%)$.

Usually, trajectories of the same class are similar to each other, which means that a certain trajectory $AT_j^c$ can be used (as a prototype) to represent the trajectory of class $c$. We define $AT_j^c$ in the following way
\begin{equation}
    \sum_{AT_i \in c}|DIS(AT_j^c,AT_i)|=\min_{AT_j \in c}\sum_{AT_i \in c}|DIS(AT_j,AT_i)|.
\end{equation}


If trajectory $AT_i$ belongs to class $c_1$, and $AT_j$ belongs to class $c_2$, then $|AT_i|\approx |AT_{i}^{c_1}|$, $|AT_j| \approx |AT_{j}^{c_2}|$. Further, we have $|AT_i\cap AT_j|\approx |AT_{i}^{c_1}\cap AT_{j}^{c_2}|$ and $|AT_{i}|- |AT_{j}|\approx|AT_{i}^{c_1}|-|AT_{j}^{c_2}|$. Thus, we have
\begin{equation}
\begin{aligned}
    &\forall{AT_{i} {\in} c}, S_{1}(AT_{i})\approx S_{1}(AT_{i}^{c}), \\
    &\sigma_{c}^{2} =  \mathbb{E}[{(S_{1}(AT_{i})-S_{1}(AT_{i}^{c}))}^{2}],
\end{aligned}
\label{eq_error}
\end{equation}
where $\sigma_{c}^{2}$ is the variance among the class $c$. This is the foundation on which random sampling works. Suppose that each extracted trajectory $AT_j$ satisfies the independent identical distribution condition (i.i.d.), then the extracted trajectories have a similar distribution as the whole. Therefore, the expectation of $S_1(AT_i)$ will not change for different sampling rates, then it can be expressed as
\begin{equation}
\begin{aligned}
    \mathbb{E}[S_1(AT_i)] &= \frac{\frac{1}{N}\sum_{j=1}^{N}(|AT_i|-|AT_j|)}
{\frac{1}{N}\sum_{j=1}^{N} |AT_i \cap AT_{j}|}\\
    &\approx \frac{\sum_{c \in \mathcal{C}}P_c(|AT_i|-|AT_{j}^{c}|)}{\sum_{c \in \mathcal{C}}P_c(|AT_i \cap AT_{j}^{c}|)} = \frac{Q_i}{M_i},
\end{aligned}
\end{equation}
where $M_i$ represents the intersection parts, $Q_i$ represents the difference parts. Now, we discuss the scale of $M_i$ and $Q_i$. 
For $M_i$, if $AT_i\in NT$, we have
\begin{equation}
    M_{i}^{NT}=\sum_{c\in \mathcal{C}} P_c |AT_{i}^{NT}\cap AT_{j}^{c}|.
\label{eq8}
\end{equation}
In Eq. (\ref{eq8}), the term $P_{NT}|AT_{i}^{NT} \cap AT_{j}^{NT}|$ plays a main role, because both $P_{NT}$ and $|AT_{i}^{NT} \cap AT_{j}^{NT}|$ are large numbers and much larger than other terms.
But when $AT_i\in GD$,
\begin{equation}
    M_{i}^{GD}=\sum_{c\in \mathcal{C}} P_c |AT_{i}^{GD}\cap AT_{j}^{c}|.
\label{eq9}
\end{equation}
No term in Eq. (\ref{eq9}) plays a main role. For example, although $|AT_{i}^{GD} \cap AT_{j}^{GD}|$ and $P_{NT}$ are large, $P_{GD}$ and $|AT_{i}^{GD} \cap AT_{j}^{NT}|$ are super small. Therefore, neither $P_{GD}|AT_{i}^{GD} \cap AT_{j}^{GD}|$ nor $P_{NT}|AT_{i}^{GD} \cap AT_{j}^{NT}|$ is large. This counteraction also happens on other terms in Eq. (\ref{eq9}). While $M_{i}^{GD}$ and $M_{i}^{GS}$ suffer more from this scenario, $M_{i}^{LD}$ and $M_{i}^{LS}$ suffer less. 
It is concluded that,
\begin{equation}
M_{i}^{NT} > M_{i}^{LD},M_{i}^{LS} > M_{i}^{GD},M_{i}^{GS}.
\label{M_i}
\end{equation}
For $Q_i$, if $AT_i\in NT$, we have
\begin{equation}
    Q_{i}^{NT}=\sum_{c\in \mathcal{C}}P_c(|AT_{i}^{NT}|-|AT_{j}^{c}|).
\end{equation}
The difference parts have following properties
\begin{equation}
\begin{aligned}
    |NT_i|-|GD_j|<&|NT_i|-|LD_j|<|NT_i|-|NT_j|, \\
    &|NT_i|-|NT_j| \approx 0, \\
    |NT_i|-|NT_j|<&|NT_i|-|LS_j|<|NT_i|-|GS_j|.\\
\end{aligned}
\label{eq_inter}
\end{equation}
For clear demonstration, we denote $AT_{i}^{NT}$ as $NT_i$ in Eq. (\ref{eq_inter}). It shows that positive and negative terms of $Q_{i}^{NT}$ cancel each other, making it close to zero.
But for $Q_{i}^{GD}$, we have
\begin{equation}
    \begin{aligned}
    |GD_i|&-|GS_j|>|GD_i|-|LS_j|>|GD_i|-|NT_j|\\
    &>|GD_i|-|LD_j|>|GD_i|-|GD_j|>0.
    \end{aligned}
\label{diff_5class}
\end{equation}
It shows that $Q_{i}^{GD}$ is a big positive value. In sum, we have 
\begin{equation}
    \begin{aligned}
        Q_{i}^{GD}&>Q_{i}^{LD}>Q_{i}^{NT}, \\
        &Q_{i}^{NT} \approx 0, \\
        Q_{i}^{NT}&>Q_{i}^{LS}>Q_{i}^{GS}.
    \end{aligned}
    \label{eqQ_i}
\end{equation}

According to Eqs. (\ref{M_i}) and (\ref{eqQ_i}), we finally obtain \begin{equation}
    \begin{aligned}
        \mathbb{E}[S_1(AT_{i}^{GS})] < &\mathbb{E}[S_1(AT_{i}^{LS})] < \mathbb{E}[S_1(AT_{i}^{NT})], \\
        &\mathbb{E}[S_1(AT_{i}^{NT})] \approx 0, \\
        \mathbb{E}[S_1(AT_{i}^{NT})] < &\mathbb{E}[S_1(AT_{i}^{LD})] < \mathbb{E}[S_1(AT_{i}^{GD})].
    \end{aligned}
    \label{eq_Es}
\end{equation}
Eq. (\ref{eq_Es}) shows that $\mathbb{E}[S_1(AT_{i}^{NT})]$ is close to 0, while other classes stay away from 0. Since the sampling satisfies IID assumption and the variance among the same class $c$ is low, $S_1(AT_{i}^{c})$ is close to $\mathbb{E}[S_1(AT_{i}^{c})]$. When $AT_i \in c$, it satisfies $S_1(AT_i)\approx S_1(AT_i^{c})$ according to Eq. (\ref{eq_error}).
These properties guarantee the right clustering of NT in most cases. Our subsequent experiments further demonstrate these properties, and locate the lowest sampling rate the model can take.

\subsection{FastATDC}

Each stage of ATDC is a transformation of a $k$-nearest neighbor algorithm. Moreover, $k\approx{N}$ in the first stage, which means it is time-consuming with the time complexity of $O(N^2)$. 
The above analysis shows that random sampling does not affect the distribution of the anomaly score when the extracted trajectories satisfy the independent identical distribution condition (i.i.d.). Therefore, we designed a FastATDC algorithm that employs a random sampling strategy at each stage of ATDC.

In the first stage, instead of traversing all trajectories in $\tau$, we only sample a small part of them randomly at a rate $r_1$. The sampled set with $r_{1}N$ trajectories is denoted by $\tau_{r_{1}} \subset \tau$, and the anomaly score can be expressed as
\begin{equation}
\label{S1-sample}
S_1(AT_i) = \frac{\sum_{j,j\neq i,j\in \tau_{r_{1}}}^{r_{1}N} \Big(|AT_i|-|AT_j|\Big)}
{\sum_{j,j\neq i,j\in \tau_{r_{1}}}^{r_{1}N} |AT_i \cap AT_j|}.
\end{equation}

With such a sampling strategy in the first stage, we reduce the computation of each trajectory from $O(N-1)$ to $O(r_{1}N)$. In fact, $r_{1}N$ could be a very small constant, i.e., $r_{1}N<10$, which can be shown in later experiments. Therefore, the time complexity of the first stage drops from $O(N^2)$ to $O(N)$, which drastically saves computation time.


After the first stage, the set of ANT $\tau_{0}$ is selected from the $N$ original trajectories. 
Then, the anomaly scores of these trajectories not in $\tau_{0}$ are calculated by Eq. (\ref{S2}). Notice that only k nearest ANT to $AT_i$ will be considered when calculating Eq. (\ref{S2}). However, there are still hundreds of ANT in $\tau_{0}$ according to the design of ATDC. Let $d$ represent the number of ANT in $\tau_{0}$. Thus the time complexity of the second stage is $O(dN+kN)$, where $N>d>k$. 

Furthermore, random sampling can be also implemented in the second stage to further reduce the time complexity.
ANT is randomly sampled at a rate $r_{2}$, and the sampled set with $r_{2}d$ trajectories is defined as $\tau_{r_2} \subset \tau_0$. Then, the $k$ nearest trajectories $\tau_{ik}$ of trajectory $AT_i$ is selected from $\tau_{r_2}$. Notice that $\tau_{ik} \subset \tau_{r_2} \subset \tau_0$. Although $r_{2}d>k$, $r_{2}d$ is now of the same order of magnitude as $k$. That means $r_{2}d\approx{mk}$, where $m<10$. Therefore, the time complexity of the second stage drops from $O(dN+kN)$ to $O(kN)$, which further reduces the computation time.

Now, the total time complexity of FastATDC is $O(kN)$, which is much lower than $O(N^2)$ when using ATDC.
The data prepossessing of FastATDC is the same as in ATDC, including popular region extracting, trajectory mapping, and trajectory indexing, as shown in Fig. 1. 
In the first stage, the anomaly score of one trajectory is calculated with Eq. (\ref{S1-sample}). In the second stage, the anomaly score $S_2(AT_i)$ is calculated using randomly selected ANT. Finally, we use a segmentation function $L$ in Eq. (\ref{L}) to distinguish between four types of abnormal trajectories (i.e., GD, LD, LS, GS) and normal trajectories.

\section{Experiments and Results}

In this section, we conduct experiments on real datasets to test the performance of FastATDC and its robustness to random sampling.

\subsection{Experimental Details}
We use six real datasets extracted from GPS trajectories of 536 cabs in the San Francisco Bay Area over 30 days\footnote{https://github.com/networkanddatasciencelab/ATDC/tree/master/Data}. The six datasets are denoted as T-1,T-2,T-3,T-4,T-5, and T-6, respectively, shown in Table \ref{tab:DATASET}. See more details about these datasets in \cite{Wang2020ATDC}. 

\begin{table}[t]
\centering
\caption{REAL DATASETS USED IN OUR EXPERIMENTS}
\label{tab:DATASET}
\renewcommand{\arraystretch}{1.15}
\begin{tabular}{cccccc}
\toprule
DS  & T & GD(\%) & LD(\%) & LS(\%) & GS(\%) \\  \midrule
T-1 & 1093 & 16 (1.5) & 24 (2.2) & 100 (9.1) & 3 (0.3) \\
T-2 & 311 & 5 (1.6) & 4 (1.3) & 15 (4.8) & 2 (0.6) \\
T-3 & 1720 & 25 (1.5) & 20 (1.2) & 73 (4.2) & 5 (0.3) \\
T-4 & 425 & 2 (0.5) & 7 (1.6) & 12 (2.8) & 2 (0.5) \\
T-5 & 1409 & 21 (1.5) & 38 (2.7) & 171 (12.1) & 13 (0.9) \\
T-6 & 1567 & 21 (1.3) & 68 (4.3) & 241 (15.4) & 25 (1.6) \\ 
\bottomrule
\end{tabular}
\end{table}

We inherit the best hyper-parameters from ATDC, i.e., $k=10$, $\mathbf{\phi}=0.04$, and $\boldsymbol{\theta}=(0.5, 0.11, -0.11, -0.5)$. In the later experiments, we optimize the parameter $\boldsymbol{\theta}$ in order to obtain better performance. We use the F1 and Macro-F1 scores to evaluate the performance of FastATDC.

\subsection{Sampling Rate Analysis}\label{AA}

The lower the sampling rate the more computation time can be saved but may impair accuracy. To locate a proper pair of sampling rates, we design two experiments to test the effect of the sampling rate on the algorithm performance and stability.

\begin{figure}[t!]
\centering
\includegraphics[width=0.48\textwidth]{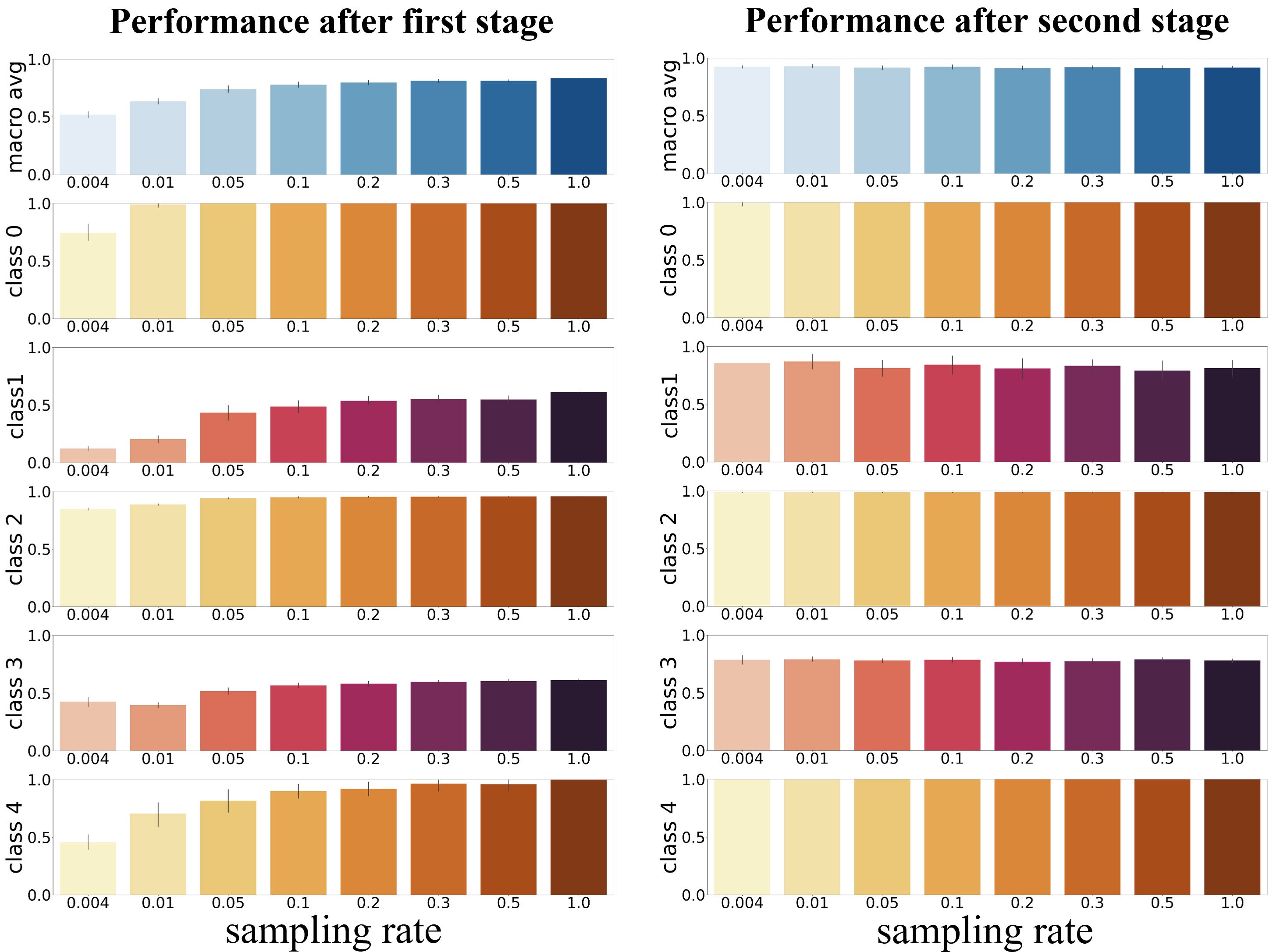}
\caption{\label{fig:First stage}Comparison between classification results of two stages when sampling is performed only in the first stage. The y-axis represents F1 scores. Particularly, the top one is the Macro-F1 scores over all classes, the others are F1 scores for each class of anomalous and normal (class 2).}
\end{figure}
\begin{figure}[t!]
\centering
\includegraphics[width=8.5cm]{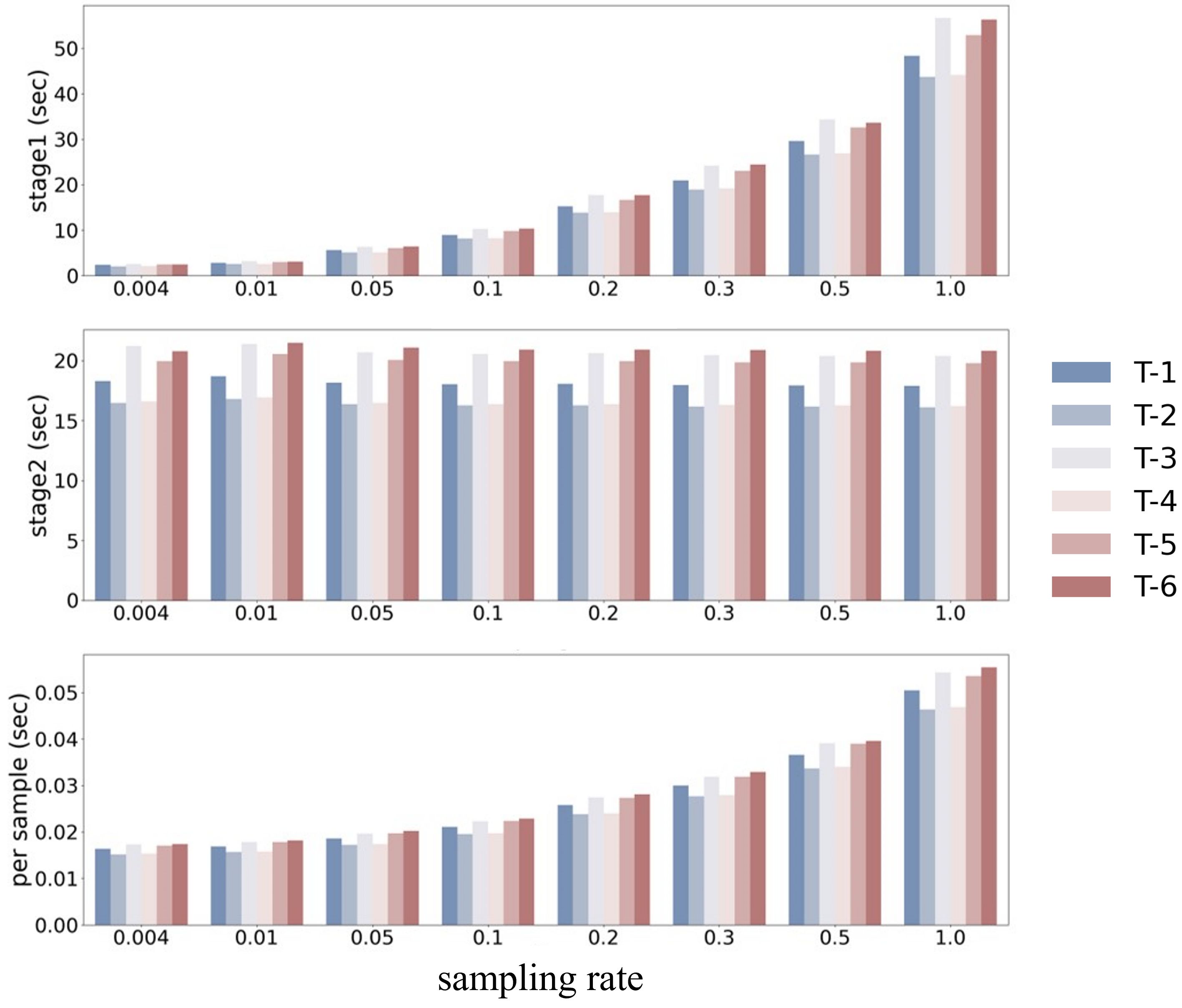}
\caption{\label{fig:Running time}Computation time when sampling only in the first stage.}
\end{figure}

\vspace{2mm}
\noindent\textbf{Sampling only in the first stage}. First, we only use random sampling at the first stage. The sampling rate is set as ${r_1}=(0.4\%, 1\%, 5\%,10\%,20\%, 30\%,50\%,100\%)$.
Note that 0.4\% is the lowest sampling rate for T-2 (the smallest dataset including 311 trajectories), which means only one trajectory is sampled. For the other datasets, only a few trajectories ($<10$) are sampled at a sampling rate of 0.4\%.

The classification results of FastATDC on T-2 are shown in Fig. 2. From above to bottom, the diagrams show Macro-F1 scores of all classes, F1 scores of class 0 (GD), class 1 (LD), class 2 (NT), class 3 (LS), and class 4 (GS), respectively. 
The result of the first stage is depicted in Fig. 2 (left), which shows that when the sampling rate drops below 10\%, the classification performance of the first stage decreases for all classes.
The result of the second stage is shown in Fig. 2 (right). Compared to the sudden drop in F1 scores in the first stage, the performance of the second stage remains high and stable. The Macro-F1 scores of all classes are still high even when the sampling rate is 0.4\%. This indicates that the second stage is significantly more important than the first one. Therefore, we can set a very low sampling rate $r_1$ in the first stage, even $r_1N < 10$.

The computation time of FastATDC with different sampling rates is shown in Fig. 3. The y-axis represents time in seconds. The first and second rows are the computation time of the first stage and second stage, respectively. The third row is the total time divided by the number of trajectories. It can be seen that the run time of the first stage is much longer compared to the second stage when there is no sampling. When the sampling rate is 0.4\%, the running time of the first stage is reduced by almost a factor of 10.

\vspace{2mm}
\noindent\textbf{Sampling in both stages}. To further shorten the computation time, we apply random sampling at both stages. 
Here, the sampling rate of the first stage is held constant at $r_1=0.4\%$. The sampling rate of the second stage is set to $r_2=\{1\%, 5\%, 10\%, 20\%, 30\%, 50\%, 70\%, 100\% \}$.

The classification results on the six datasets are shown in Fig. 4. From the top row of Fig. 4., the Marco-F1 score begins to drop when the sampling rate $r_2$ is below 30\%. From the diagrams of each class, it can be seen that three classes (LD, LS, GS) are more vulnerable to a low sampling rate, while the other two classes (GD, NT) are more immune to a low sampling rate. The reason why a higher rate is required in the second stage is that the variances (in Eq. (\ref{eq_error})) among class LD and LS are large but their prototype are not distinct enough from the prototype of $NT(NT \notin ANT)$, which causes overlapping.
Notice that in ATDC, experiments have been conducted on different intervals $[-\phi,\phi]$ to select the absolute normal trajectories. The conclusion is that neither a large $\phi$ nor a small one draws good performance. Therefore, we can not shorten the computation time via a tighter interval $[-\phi,\phi]$ without affecting accuracy. Our experiment shows that, apart from setting a tighter interval, random sampling of ANT in the interval $[-\phi,\phi]$ can reduce computation time without hindering classification accuracy. 

In a word, we can utilize random sampling at both stages, and set the sampling rate to 0.4\% and 30\% for the first and the second stage, respectively.

\begin{figure}[t!]
\centering
\includegraphics[width=0.48\textwidth]{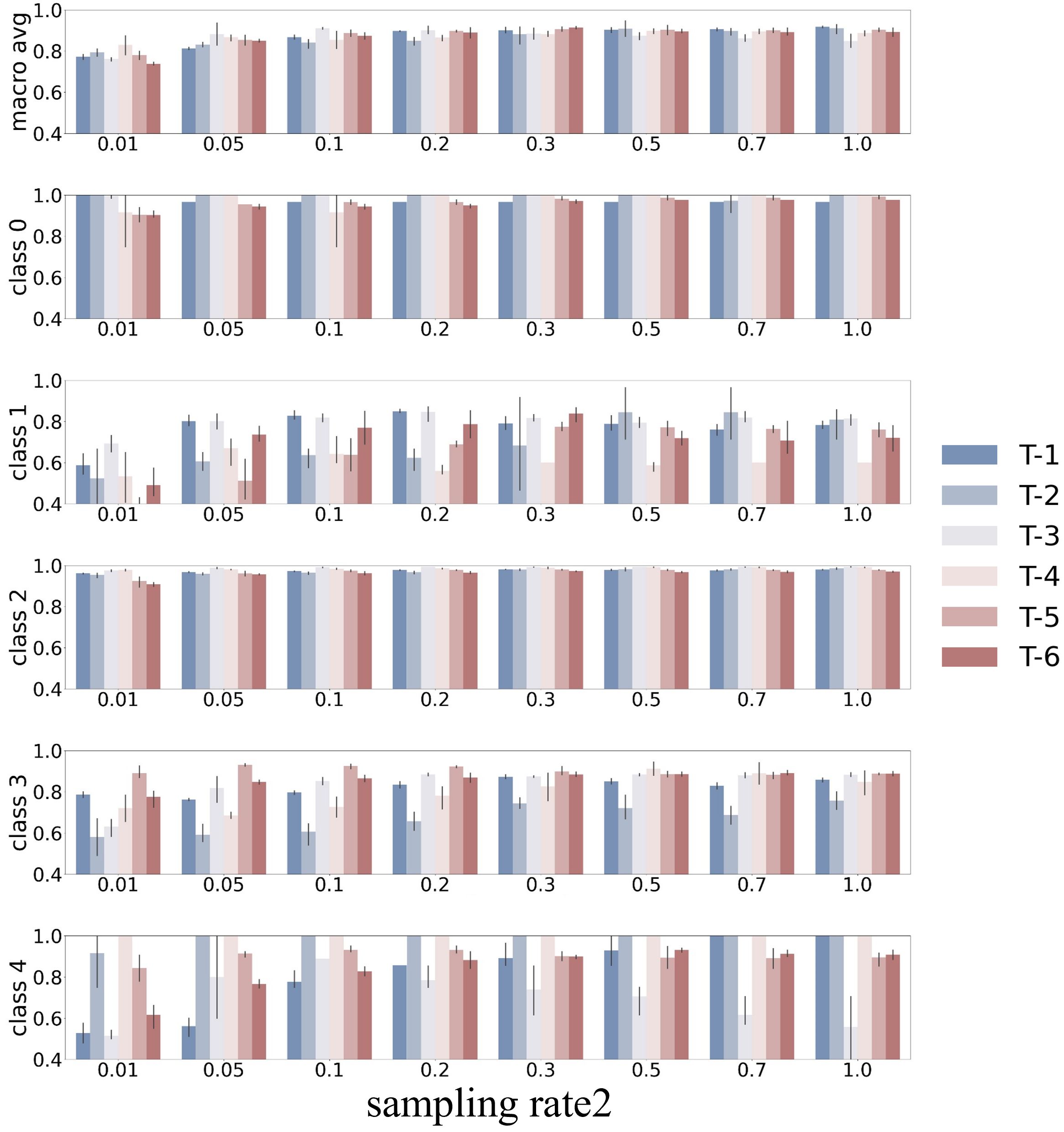}
\caption{\label{fig:Second stage}The classification results of the second stage when sampling is performed in both stages.}
\end{figure}

\begin{table}[t]
\caption{Performance comparison between FastATDC and ATDC.}
\label{tab:f1-comparison}
\centering
\setlength{\tabcolsep}{5.6pt}
\renewcommand{\arraystretch}{1.2}
\begin{tabular}{ccccccc}
\toprule
DS             & Method  & GD     & LD     & LS     & GS     & Macro-F1 \\ 
\midrule
\multirow{2}{*}{T-1} & ATDC     & 0.9677 & 0.8095 & 0.8    & 1      & 0.8943 \\
                     & FastATDC & 0.9677 & 0.8026 & 0.8488 & 0.9179 & 0.8842 \\ \hline
\multirow{2}{*}{T-2} & ATDC     & 1      & 0.8571 & 0.7778 & 1      & 0.9087 \\
                     & FastATDC & 1      & 0.8476 & 0.7326 & 1      & 0.8951 \\ \hline
\multirow{2}{*}{T-3} & ATDC     & 1      & 0.8    & 0.8554 & 0.8889 & 0.8861 \\
                     & FastATDC & 1      & 0.7378 & 0.8888 & 0.8472 & 0.8685 \\ \hline
\multirow{2}{*}{T-4} & ATDC     & 1      & 0.8333 & 0.7857 & 1      & 0.9048 \\
                     & FastATDC & 1      & 0.7665 & 0.8230 & 1      & 0.8974 \\ \hline
\multirow{2}{*}{T-5} & ATDC     & 1      & 0.7473 & 0.9706 & 0.9296 & 0.9086 \\
                     & FastATDC & 0.9816 & 0.7665 & 0.9207 & 0.9292 & 0.8995 \\ \hline
\multirow{2}{*}{T-6} & ATDC     & 0.9767 & 0.8767 & 0.9144 & 0.9411 & 0.9273 \\
                     & FastATDC & 0.9634 & 0.8627 & 0.8926 & 0.8924 & 0.9028 \\ \bottomrule
\end{tabular}
\end{table}

\begin{table}[t!]
\caption{The compuation time comparison between FastATDC and ATDC. Recorded values are in seconds per 100 trajectories.}
\label{tab:time-compare}
\centering
\setlength{\tabcolsep}{6pt}
\renewcommand{\arraystretch}{1.1}
\begin{tabular}{cccc}
\toprule
Dataset & ATDC   & FastATDC & speedup         \\ \midrule
T-1       & 9.011  & 0.456    & 20$\times$      \\
T-2       & 3.102  & 0.308    & 10$\times$       \\
T-3       & 11.61  & 0.591    & 20$\times$      \\
T-4       & 4.106  & 0.309    & 13$\times$      \\
T-5       & 10.60  & 0.679    & 16$\times$      \\
T-6       & 9.735  & 0.688    & 14$\times$      \\ 
\bottomrule
\end{tabular}
\end{table}

\subsection{Comparative Evaluation}

Experiments on real datasets are performed to compare the performance of FastATDC with other baselines. Since random sampling influence the distribution of anomaly score slightly, we optimize $\boldsymbol{\theta}$ for different datasets. $\theta_1$ and $\theta_4$ are the same as before, which are 0.5 and -0.5. $\theta_2$ is set to 0.1 for T-1 and T-3, 0.11 for T-2 and T-5, 0.075 for T-4, and 0.09 for T-6. $\theta_3$ is set to -0.11 for T-1 and T-3, -0.13 for T-2 and T-5, -0.085 for T-4, and -0.135 for T-6.

\vspace{2mm}
\noindent\textbf{FastATDC vs. ATDC}.
First, we compare FastATDC with ATDC in the six real datasets. The results are shown in Table \ref{tab:f1-comparison}.
Although we apply a random sampling strategy in both stages, the Macro-F1 scores on all datasets are still very high. There is no significant drop in classification accuracy over all classes. This shows that FastATDC has an excellent performance in the detection and classification of anomalous trajectories.
Table \ref{tab:time-compare} shows the running time per 100 trajectories on average. It can be seen that FastATDC is 10 to 20 times faster than ATDC.

\vspace{2mm}
\noindent\textbf{FastATDC vs. baselines}.
Second, we compare FastATDC with other baseline algorithms (iBAT \cite{Zhang2011iBAT} and Density). Here, we do not regard the classes of anomalous trajectories and only consider “normal” versus “anomalous”. Firstly, we regard all four kinds of anomalous trajectories as the same class, i.e., anomalous trajectories (Case 1). Secondly, we consider another extreme case, that is, all global anomalies (GD and GS) are regarded as anomalous trajectories, and local anomalies (LD and LS) are regarded as normal trajectories (Case 2). As shown in Tables \ref{tab:Case1} and \ref{tab:Case2}, FastATDC outperforms other methods on datasets T-1, T-2, T-3, T-4 in Case 1, and on datasets T-1, T-3, T-5, T-6 in Case 2. These results show that the FastATDC algorithm excels in the baseline algorithms by a large margin and even outperforms ATDC in some cases.

\begin{table}[t]
\caption{The results of four algorithms in six datasets (Case 1). Best values are printed in bold.}
\label{tab:Case1}
\centering
\setlength{\tabcolsep}{6pt}
\renewcommand{\arraystretch}{1.1}
\begin{tabular}{ccccc}
\toprule
Dataset & FastATDC & ATDC   & iBAT   & Density \\ \midrule
T-1      & \textbf{0.9216}    & 0.8621 & 0.7483 & 0.7063  \\
T-2      & \textbf{0.9134}    & 0.8077 & 0.8846 & 0.7692  \\
T-3      & \textbf{0.9257}    & 0.9136 & 0.8374 & 0.7642  \\
T-4      & \textbf{0.9054}    & 0.9048 & 0.6522 & 0.5217  \\
T-5      & 0.9106    & \textbf{0.9283} & 0.7737 & 0.5638  \\
T-6      & 0.9146    & \textbf{0.9275} & 0.7746 & 0.5915  \\ \bottomrule
\end{tabular}
\end{table}

\begin{table}[t]
\caption{The results of four algorithms in six datasets (Case 2). Best values are printed in bold.}
\label{tab:Case2}
\centering
\setlength{\tabcolsep}{6pt}
\renewcommand{\arraystretch}{1.1}
\begin{tabular}{ccccc}
\toprule
Dataset & FastATDC & ATDC   & iBAT   & Density \\ \midrule
T-1      & \textbf{0.9746}   & 0.9474 & 0.8947 & 0.9474  \\
T-2      & 0.9873   & \textbf{1}      & 0.7143 & \textbf{1}       \\
T-3      & \textbf{0.9881}   & 0.9831 & 0.8333 & 0.9     \\
T-4      & 0.9907   & \textbf{1}      & 0.5    & \textbf{1}       \\
T-5      & \textbf{0.9810}   & 0.9697 & 0.7353 & 0.8529  \\
T-6      & \textbf{0.9837}   & 0.9574 & 0.6522 & 0.8478  \\
\bottomrule
\end{tabular}
\end{table}

\section{Conclusion and Future work}
This paper analyzed the limitations of the ATDC algorithm and proposed an improved ATDC algorithm called FastATDC, which obtains superior performance with lower time complexity. Experiments on real datasets show that FastATDC is 10 to 20 times faster than ATDC, and outperforms other baseline algorithms. The excellent performance of FastATDC shows that it is promising to be used in various ITS applications.
A limitation of FastATDC is that the DIS distance is unable to measure the difference between an ongoing partition of trajectory and complete trajectories in the dataset. Therefore, online detection and classification of anomalous trajectories will be studied in future work.





\section*{Acknowledgment}
We thank Dr. Wei Qian for her valuable suggestions on our work and for her review of our mathematical formulas.





\bibliographystyle{IEEEtran}
\bibliography{main}


\end{document}